\documentclass[11pt,a4paper]{article}
\usepackage[hyperref]{acl2018}
\usepackage{amsmath}
\usepackage[linesnumbered,algoruled,boxed,lined]{algorithm2e}
\usepackage{algorithmic}
\usepackage{latexsym}
\usepackage{multirow}
\usepackage{array,graphicx}
\usepackage{url}
\usepackage{microtype}
\usepackage{enumitem}
\usepackage{hyperref}
\usepackage{floatrow}
\usepackage{caption}
\usepackage{wrapfig}
\usepackage{xcolor}
\usepackage{color}
\usepackage{changes}

\definechangesauthor[color=purple]{PC}
\definechangesauthor[color=red]{CC}

\aclfinalcopy 

\title{The importance of fillers for text representations of speech transcripts}  

\author{Tanvi Dinkar\textsuperscript{\rm 1}\footnote[1]{Equal contribution}, Pierre Colombo\textsuperscript{\rm 1,2}\thanks{\, Equal contribution},\\ \textbf{Matthieu Labeau\textsuperscript{\rm 1}, Chlo\'e Clavel\textsuperscript{\rm 1}}\\
\textsuperscript{\rm 1}LTCI, Telecom Paris, Institut Polytechnique de Paris, \textsuperscript{\rm 2}IBM GBS France\\
\textsuperscript{\rm 1}firstname.lastname@telecom-paris.fr
}

\date{}

\begin{document}
\maketitle
\begin{abstract}
     
     While being an essential component of spoken language, fillers (e.g. ``um" or ``uh") often remain overlooked in Spoken Language Understanding (SLU) tasks. We explore the possibility of representing them with deep contextualised embeddings, showing improvements on modelling spoken language and two downstream tasks --- predicting a speaker's stance and expressed confidence.  
     
     
     
\end{abstract}

\section{Introduction}
Disfluencies are interruptions in the regular flow of speech, such as pausing silently, repeating words, or interrupting oneself to correct something said previously \cite{Disfluency}.
They commonly occur in spoken language, as spoken language is rarely fluent. \textit{Fillers} are a type of disfluency that can be a sound (``um" or ``uh") filling a pause in an utterance or conversation.

Recent work has shown that contextualised embeddings pre-trained on large written corpora can be fine-tuned on smaller spoken language corpora to learn structures of spoken language \cite{Tran2019}. However, for NLP tasks, fillers and all disfluencies are typically removed in pre-processing, as NLP models achieve highest accuracy on syntactically correct utterances. This contradicts linguistic studies, which show that fillers are an essential and informative part of spoken language \cite{Clark2002_using,Yoshida2010_disfluency,Brennan1995_feeling,corley2007, stolcke1996statistical}. 

So far, the information carried by fillers has only been studied using hand crafted features, for example in \citet{le2017and,Saini2017_disfluency,FOK}. Besides, \citet{valentin2017} show that pre-trained word embeddings such as Word2vec \cite{word2vec}, have poor representation of spontaneous speech words such as ``uh", as they are trained on written text and do not carry the same meaning as when used in speech. 
We address the matter of representing fillers with deep contextualised word representations \cite{devlin2018bert}, and investigate their usefulness in NLP tasks for spoken language, without handcrafting features. 

Hence, the present work is motivated by the following observations: (1) Fillers play an important role in spoken language. For example, a speaker can use fillers to inform the listener about the linguistic structure of their utterance, such as in their (difficulties of) selection of appropriate vocabulary while informing the listener about a pause in their upcoming speech stream \cite{Clark2002_using}.
(2) Fillers and prosodic cues have also been linked to a speaker's \textit{Feeling of Knowing (FOK)} or \textit{expressed confidence}, that is, a speaker's certainty or commitment to a statement \cite{smith1993_course}.  \citet{Brennan1995_feeling} observed that fillers and prosodic cues contribute to the listener's perception of the speaker's expressed confidence in their utterance, which they refer to as the \textit{Feeling of Another's Knowing (FOAK)}, also observed by \cite{wollermann2013disfluencies}. (3) Recent work has shown that fillers have been successful in \textit{stance} prediction (stance referring to the subjective spoken attitude towards something) \cite{le2017and}.

\textbf{Aim of this work:} 
We want to verify that these observations are still valid when we represent fillers in an automatic and efficient way. Hence, our contributions are as follows: (1) Fillers contain useful information that can be leveraged by deep contextualised embeddings to better model spoken language and thus should not be removed. In addition, we study which filler representation strategies are best suited to our task of Spoken Language Modelling (SLM) and investigate the learnt positional distribution of fillers. (2) We show that in a spontaneous speech corpus of spoken monologues, fillers are a discriminative feature in predicting the perception of expressed confidence of the speaker, and perception of a speaker's stance (which we measure by sentiment). \section{Models and Data description}
\subsection{Model Description}
For our work, we consider the two fillers ``uh" and ``um" (see \autoref{ssec:data_description}).  To obtain contextualised word embeddings for fillers, we use bidirectional encoder representations from transformers (BERT) \cite{devlin2018bert}, as it has achieved SOTA performance on several NLP benchmarks and are better than Word2Vec for word sense disambiguation by integrating context \cite{BartunovKOV15}.


\begin{table*}[!htbp]
    \centering
\resizebox{\textwidth}{!}{\begin{tabular}{c|c} \hline 
Token. & Output Tokenizer \\
\hline
   Raw &  (umm) Things that (uhh)  you usually wouldn't find funny were in this movie. \\
   $\mathcal{T}_1$ & [`umm', `things', `that', `uh', `you', `usually', `wouldn', ``'", `t', `find', `funny', `were', `in', `this', `movie', `.'] \\
      $\mathcal{T}_2$ & [`[$FILLER_{UMM}$]', `things', `that', `[$FILLER_{UHH}$]', `you', `usually', `wouldn', ``'", `t', `find', `funny', `were', `in', `this', `movie', `.'] \\
         $\mathcal{T}_3$ & [`[FILLER]', `things', `that', `[FILLER]', `you', `usually', `wouldn', ``'", `t', `find', `funny', `were', `in', `this', `movie', `.'] 
         
\end{tabular}}
\caption{Filler representation using different token representation strategies}
\label{table:tokenizer_strat}\vspace{-0.2cm}
\end{table*}
\subsubsection{Spoken Language Modelling}
\label{description SLM}

For SLM, we use the masked language modelling objective (MLM). It consists of masking some words of the input tokens at random, and then predicting these masked tokens. The MLM objective is classically used to pre-train and then fine-tune BERT. Here, we use this MLM objective to fine-tune a pretrained BERT on a spoken language corpus (see \autoref{ssec:data_description}). Each experiment requires a token representation strategy $\mathcal{T}_i$ and a pre-processing strategy $\mathcal{P}_{Si}$ (additional details are given in the \autoref{alg:alg1} in Supplementary). 


The \textbf{token representation strategies} are particularly important for our task, for BERT to learn the distribution of fillers. The three token representation strategies $(\mathcal{T}_1,\mathcal{T}_2,\mathcal{T}_3)$, are described as follows: In $\mathcal{T}_1$, no special treatment is done to the fillers\footnote{It is interesting to note that BERT provides embedding for ``uh" or ``um" despite being trained on written text (Wikipedia, BooksCorpus \cite{book}, Word Benchmark \cite{word_bench}.}, i.e BERT will use its a priori knowledge of the fillers ``uh" or ``um" to model the language. In $\mathcal{T}_2$, ``uh" and ``um" are distinguished from other tokens by a special filler tag, and are represented as two different tokens respectively; this strategy aims at forcing BERT to learn a new embedding that focuses both on the position and the context of the fillers. In $\mathcal{T}_3$, 
both fillers are represented as the same token, suggesting that they have the same pragmatic meaning and are interchangeable. A concrete example is given in \autoref{table:tokenizer_strat}. 

\textbf{Pre-processing strategies}, $(\mathcal{P}_{S1},\mathcal{P}_{S2},\mathcal{P}_{S3})$, are as follows: In $\mathcal{P}_{S1}$, the sentences have all fillers removed, both during training and inference. In $\mathcal{P}_{S2}$, the sentences have the fillers kept during training, but are removed at inference. In $\mathcal{P}_{S3}$, the fillers are kept both during training and inference. For each pre-processing and token representation strategy, we optionally fine-tune BERT using the same Masked Language Model (MLM) objective as in the original paper \cite{devlin2018bert}. Note, if we do not fine-tune, the training dataset ($\mathcal{D}_{train}$) is not used and therefore $\mathcal{P}_{S1}$ and $\mathcal{P}_{S2}$ are equivalent. For language modelling we report the perplexity (ppl) measure to evaluate the quality of the model.

\subsubsection{Confidence and Sentiment Prediction}
\label{ssec:fk_pred}

In both our confidence prediction and sentiment analysis task, our goal is to predict a label of confidence/sentiment using our BERT text representations that include fillers.  Formally, our confidence/sentiment predictor is obtained by adding a Multi-Layer Perceptron (MLP) on top of a BERT, which has been optionally fine-tuned using the MLM. The MLP is trained by minimising the mean squared error (MSE) loss (additional details are given in \autoref{alg:alg2} in Supplementary). We keep the same token representation and pre-processing strategies from Section \ref{description SLM}. 



\subsection{Data Description}\label{ssec:data_description}
We use the Persuasive Opinion Mining (POM) dataset \cite{park2014computational}, a dataset of 1000 English monologue videos. Speakers recorded themselves giving a movie review, freely available on ExpoTV.com. The movies were rated from 1 star (most negative) to 5 stars (most positive). Annotators were asked to label the video for high-level attributes. For confidence, annotators (3 per video) were asked ``How confident was the reviewer?", and had to each give a label respectively; from 1 (not confident) to 7 (very confident), after watching the entire review. 
Similarly for sentiment, the annotators were asked ``How would you rate the sentiment expressed by the reviewer towards this movie?", and were asked to give a label from 1 (strongly negative) to 7 (strongly positive).

We choose this dataset for the following reasons: 
(1) The corpus has been manually transcribed with fillers ``uh" and ``um", where $ \approx 4\%$ of the speech consists of fillers (for comparison, the Switchboard \cite{godfrey1992switchboard} dataset of human-human dialogues, consists of $\approx 1.6\%$ of fillers \cite{Shriberg2001_errrr}). Sentence markers have been manually transcribed, with the practice of the filler being annotated sentence-initially, if the filler occurs between sentences. (2) The dataset consists of monologues, where the speaker is conscious of an \textit{unseen} listener, but dialogue-related disfluencies (such as backchannels) are not present, allowing us to concentrate on fillers of the narratives of the speaker \cite{swerts1998filled}. (3) Only reviews with a 1-2 star or a 5 star rating were chosen for annotation, to clearly demarcate sentiment/stance polarity. (4) \textit{FOAK}, which we measure by the given label of confidence, has been annotated with high inter-annotator agreement (Krippendorff’s alpha = 0.73). 

Details can be found in the supplementary material and in \citet{park2014computational}.
Confidence labels are obtained by taking the root mean square (RMS) value of the labels given by the 3 annotators\footnote{Though the inter-annotator agreement for confidence is high, we choose RMS as a way to handle disagreement between annotators. For example, annotation labels $\{3,5,7\}$ would result in mean value of $5$, not highlighting that one annotator found the reviewer particularly confident. The RMS value however ($\approx 5.3$), slightly enhances the high confidence label.}. Sentiment labels are calculated by taking the mean of the 3 labels, which were obtained from \citet{sentiment_labels}\footnote{A toolkit for multimodal analysis. Please refer to the \textbf{Usage} and the \textbf{Supported Datasets}
sections, which include instructions to download the data.}.

\section{Experiments and Analysis}
\label{exp}

\begin{table*}
    \centering
\begin{tabular}{c|ccccc|cccc|ccc} \cline{1-4} \cline{6-8} \cline{10-13}
   Fine. &Setting & Token. & Ppl & & Setting & Token. & Ppl & &   Fine.  & Model & FOAK & Sent \\ \cline{1-4} \cline{6-8} \cline{10-13}
    
 \multirow{3}{*}{{ w/o}}    & $\mathcal{P}_{S1}$& $\mathcal{T}_1$  & 22 & &\multirow{2}{*}{$\mathcal{P}_{S3}$} &\multirow{2}{*}{$\mathcal{T}_1$ } & \multirow{2}{*}{\textbf{4.6}}&    &\multirow{3}{*}{w/o} &$ \mathcal{P}_{S1}$ & 1.47 & 1.98\\
 & $\mathcal{P}_{S2}$ & $\mathcal{T}_1$  & 22 & & & & & & &$ \mathcal{P}_{S2}$ & 1.45 & 1.75 \\
   & $\mathcal{P}_{S3}$& $\mathcal{T}_1$  & \textbf{20}& &\multirow{2}{*}{$\mathcal{P}_{S3}$} &\multirow{2}{*}{$\mathcal{T}_2$ } & \multirow{2}{*}{4.7}& & &$ \mathcal{P}_{S3}$ & \textbf{1.30} & \textbf{1.44}\\ \cline{1-4}  \cline{10-13}
    \multirow{3}{*}{{ w}}  &$ \mathcal{P}_{S1} $& $\mathcal{T}_1$  & 5.5 & & & & &   & \multirow{3}{*}{w} &$ \mathcal{P}_{S1}$ &1.32 & 1.39\\
    & $ \mathcal{P}_{S2}$& $\mathcal{T}_1$  & 5.6 & &\multirow{2}{*}{$\mathcal{P}_{S3}$} & \multirow{2}{*}{$\mathcal{T}_3$ }&\multirow{2}{*}{4.7} & & &$ \mathcal{P}_{S2}$ &1.31 & 1.40\\
       & $\mathcal{P}_{S3}$& $\mathcal{T}_1$ & \textbf{4.6} & & & & & & &$ \mathcal{P}_{S3}$ & \textbf{1.24} &\textbf{1.22} \\\cline{1-4} \cline{6-8} \cline{10-13}
        \multicolumn{3}{c}{(a)}& &&\multicolumn{2}{c}{(b)}   & & &\multicolumn{3}{c}{(c)}  
\end{tabular}
\caption{From left to right, the (a) LM Task, (b) Best token representation, (c) MSE of Confidence (FOAK) and the Sentiment (Sent) prediction task. Wilcoxon test (10 runs with different seeds) has been performed. Highlighted results exhibit significant differences ($\text{p-value} < 0.005$). Data split is fixed according to \citet{SDK} and results are given on the test set (see \autoref{sec:Suplement}) for additional details} 
\label{table:SLM}\vspace{-0.3cm}
\end{table*}

\subsection{Information contained by fillers can be leveraged to model spoken language.}\label{ssec:perplexity_language}
\textbf{Language Modelling with fillers. } We compare the perplexity of the LM with different pre-processing strategies with a fixed token representation $\mathcal{T}_1$. Results are reported in \autoref{table:SLM}(a). We compare $\mathcal{P}_{S1}$,$\mathcal{P}_{S2}$ $\mathcal{P}_{S3}$ with or without fine-tuning and observe that adding fillers, both during training and inference, leads to a model with lower perplexity and a perplexity reduction of at least $10\%$. Hence, fillers contain information that can be leveraged by BERT.

As shown, the fine-tuning procedure reduces the perplexity of the language model. Even without fine-tuning, we observe that $\mathcal{P}_{S3}$ outperforms $\mathcal{P}_{S1}$/$\mathcal{P}_{S2}$, as the perplexity reduces when adding fillers. This suggests that BERT has a priori knowledge of spoken language, in terms of fillers. 

Hence, fillers can be leveraged to reduce uncertainty of BERT for SLM. This is not an expected result, as intuitively, one might think that the perplexity would reduce when fillers are excluded from both training and inference, due to the fact that the utterance is shorter and ``simplified". The fact that $\mathcal{P}_{S3}$ outperforms the other pre-processing methods also suggests that the MLM procedure is an effective way to learn this information. \indent


\textbf{Best Token representation:} We observe that $\mathcal{T}_1$ outperforms the other representations in a fine-tuning setting, as shown in \autoref{table:SLM}(b). Given the restricted size of our data and the dimension of the BERT embeddings (768), it is better to keep the existing representations (with $\mathcal{T}_1$), than adding and learning new representations from scratch.
Interestingly, $\mathcal{T}_2$ and $\mathcal{T}_3$ perform the same. This can be explained by ``um" and ``uh" being only distinguished in duration \cite{Clark2002_using}, the hypothesis being that ``uh" is used for a shorter pause in speech; which cannot not be reflected in text. Given these results, we fix $\mathcal{T}_1$ as the token representation strategy for the rest of the experiments.

\begin{figure}[h]
\begin{center}
\includegraphics[width=\columnwidth,height=1.7cm]{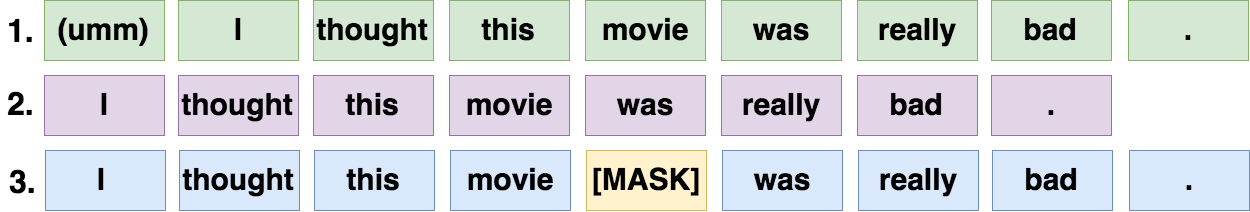} 
\caption{Predicting the probability of a filler, where 1. Raw input, 2. Pre-processed text with the filler removed, and 3. Illustrates the [MASK] procedure for predicting the probability of a filler at position 5.}\label{fig:pos_fillers}
\end{center}
\end{figure}

\begin{figure}
\begin{center}
\includegraphics[scale=0.20]{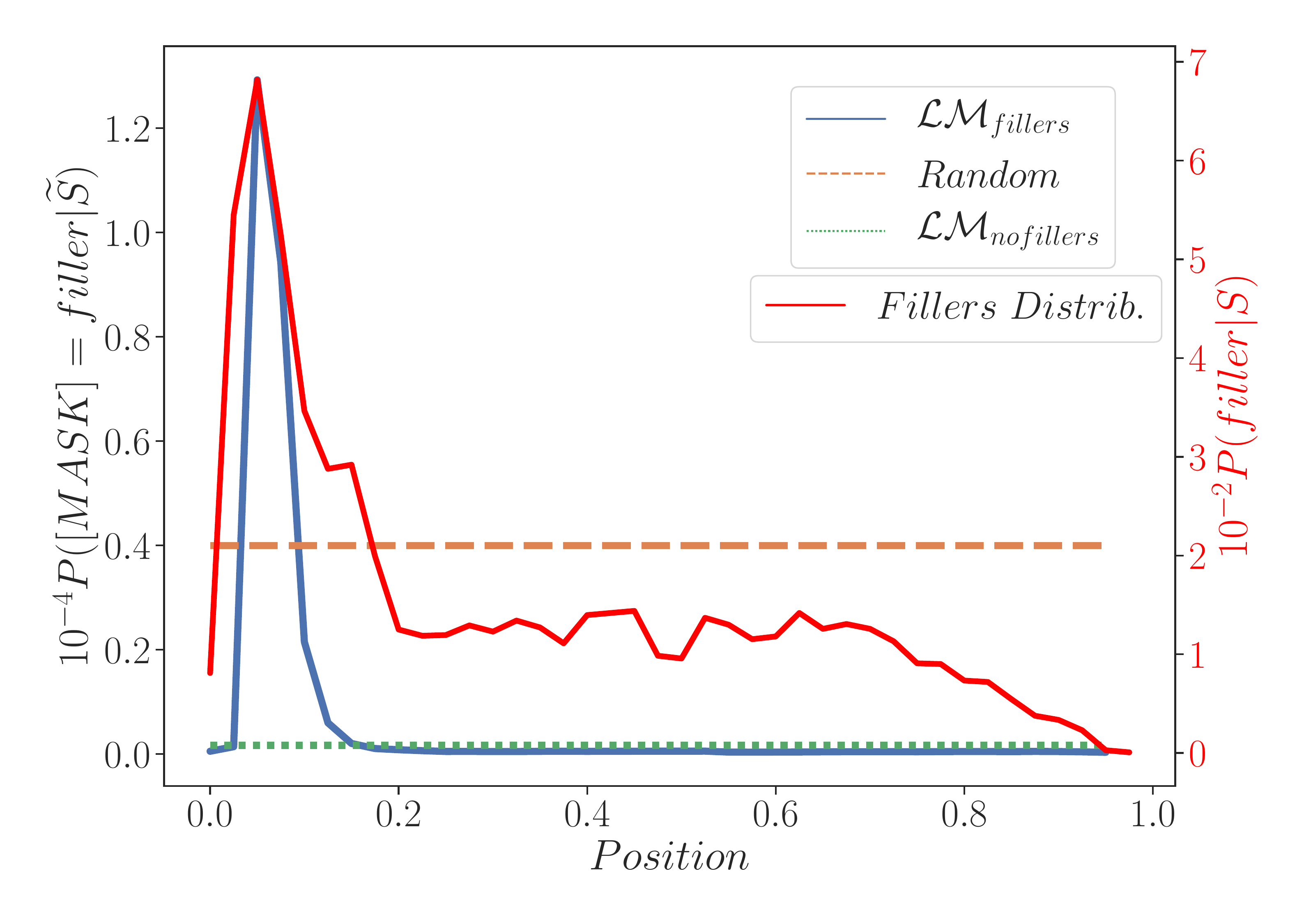}
\caption{Predicting the position of fillers. \textit{Fillers Distrib.} stands for the actual filler distribution in the dataset.  \textit{Random} stands for the random predictor which predicts $P([MASK] = filler|\widetilde{S}) = \frac{2}{|\mathcal{V}|}$ where $|\mathcal{V}|$ is the size of the vocabulary, and 2 represents both fillers.\vspace{-0.4pt}}
\label{fig:filler_position}
\end{center}
\end{figure}

\textbf{Learnt Positional distribution of fillers:} We additionally test whether our model has learnt information about the placement of fillers. We use fine-tuned BERT on $\mathcal{D}_{train}$ with fillers to see where the model estimates the most probable position of the fillers (which we call $\mathcal{LM}_{fillers}$) to be. Given a sentence $S$ of length $L$, we insert after word $j$ the mask token (`[MASK]') to obtain the corrupted sentence $\widetilde{S}$\footnote{For clarity we abuse the notation and remove dependence in $j$.}. We compute the probability of the appearance of a filler in position $j+1$ according to the LM, which corresponds to $P([MASK] = filler|\widetilde{S})$, as illustrated by \autoref{fig:pos_fillers}. 
Formally, we plot the average of the probability of the masked word to be a filler given its position in the sentence, as shown in \autoref{fig:filler_position}. We observe that the fine-tuned BERT on $\mathcal{D}_{train}$ with fillers ($\mathcal{LM}_{fillers}$) predicts with high probability  fillers occurring at the first position in the sentence (please refer to \autoref{table:sample} supplementary for example sentences).  This is consistent with the actual distribution of fillers in the dataset, as can be seen in \autoref{fig:filler_position}. The fine-tuned BERT on $\mathcal{D}_{train}$ without fillers ($\mathcal{LM}_{no fillers}$) predicts a constant low probability. Given the available segmentation of sentence boundaries (fine-grained discourse annotations are not available), it is interesting to note that our model was able to capture similar positional distribution of fillers that are reported in \citet{swerts1998filled,Shriberg2001_errrr,swerts1994prosody,Yoshida2010_disfluency}. 

In this section we show that although BERT uses contextualised word embeddings, the information contained in fillers can be leveraged to achieve a better modelling of spoken language.



 
\vspace{-0.4pt}


\subsection{Fillers are a discriminative feature for \textit{FOAK} and \textit{stance} prediction.}
 We observe the impact that fillers have on two downstream tasks, a novel FOAK prediction task, and a ubiquitous sentiment analysis task. 
Psycholinguistic studies have observed the link between fillers and expressed confidence \cite{smith1993_course,Brennan1995_feeling,wollermann2013disfluencies}. Previous research on the link between fillers and their relation to a speaker's expressed confidence has been confined to a narrow range of QA tasks \cite{schrank2015automatic}. Fillers have also been linked to stance prediction \cite{le2017and}, which we measure using sentiment. We show that in a spontaneous speech corpus of spoken monologues, fillers can play a role in predicting both the perception of the speaker's expressed confidence and 
speaker's stance. 

In \autoref{table:SLM}(c) we observe that both with and without fine-tuning the $\mathcal{P}_{S3}$ decreases the MSE compared to $\mathcal{P}_{S1}$ and $\mathcal{P}_{S2}$. $\mathcal{P}_{S1}$ and $\mathcal{P}_{S2}$ have similar MSE because fillers are not added during the inference phase. We observe that $\mathcal{P}_{S2}$ leads to higher MSE, possibly because of the discrepancy created between $\mathcal{D}_{train}^{labelled}$ and $\mathcal{D}_{test}^{labelled}$. This shows that fillers can be a discriminative feature in both FOAK and stance \cite{le2017and} prediction, apart from overt lexical cues \footnote{by overt lexical cues, we mean words that explicitly express uncertainty/confidence, such as \textit{maybe, I'm unsure} or sentiment, \textit{amazing, disgusting})}. 

\textbf{Does the addition of fillers always improve the results for downstream spoken language tasks?}
In the \autoref{ssec:perplexity_language}, we show that by including fillers , the MLM achieves a lower perplexity. An assumption one could make based on the work by \citet{radford2019language}, is that with this model, the results for any further downstream task would be improved by the presence of fillers. However, we observe that to predict the persuasiveness of the speaker (using the high level attribute of persuasiveness annotated in the dataset \cite{park2014computational}), following the same procedure as outlined in \autoref{ssec:fk_pred}, that fillers, in fact, are not a discriminative feature.  



\section{Conclusion}
\label{conclusion}

 
When working with deep contextualised representations of transcribed spoken language, we showed that retaining fillers can improve results, both when modelling language and on
a downstream task (FOAK and stance prediction). Besides, we propose and compare various token representation and pre-processing strategies in order to integrate fillers. We plan to extend these results by studying the mixing of such textual filler-oriented representations with acoustic representations, and further investigate the representation of fillers learnt during pre-training.

\section*{Acknowledgements}
This project has received funding from the European Union's Horizon 2020 research and innovation programme under grant agreement No 765955 and the French National Research Agency's grant ANR-17-MAOI.



\bibliography{acl2018}
\bibliographystyle{acl_natbib}

\clearpage
\section{Supplementary} \label{sec:Suplement}
\label{supplementary}
\begin{table*}[!h]
    \centering
\resizebox{\textwidth}{!}{\begin{tabular}{c|c} \hline 
Token. & Output Tokenizer \\
\hline
   Raw & (umm) It's an interesting movie to say the least.  \\
   T1 & [`umm', `it', ``'", `s', `an', `interesting', `movie', `to', `say', `the', `least', `.'] \\ 
   T2 & [`[$FILLER_{UMM}$]', `it', ``'", `s', `an', `interesting', `movie', `to', `say', `the', `least', `.'] \\
      T3 & [`[FILLER]', `it', ``'", `s', `an', `interesting', `movie', `to', `say', `the', `least', `.'] \\
      
         
\end{tabular}}
\caption{Additional example of the different token representation strategies}
\label{table:tokenizer_strat_additional}
\end{table*}

\subsection{Model}
\textbf{Detailed algorithms}: In \autoref{alg:alg1} and \autoref{alg:alg2}, we provide additional details of the procedure used for the language modelling task and confidence prediction task. For stance prediction, the procedure is the same as for confidence. 

\begin{algorithm}
 \SetKwInOut{Input}{Input}
\SetKwInOut{Output}{Output}
\Input{ $\mathcal{P}_{Si}$, $\mathcal{T}_i$, Pret. BERT $\mathcal{LM}$}
\Output{$(\mathcal{LM},Perplexity)$}
$(\mathcal{D}_{train}, \mathcal{D}_{dev}, \mathcal{D}_{test}) \leftarrow$ (train, dev, test set) according to ($\mathcal{P}_{Si}$,$\mathcal{T}_i$)

  \If{Do Finetuning}{
   $\mathcal{LM} \leftarrow \mathcal{LM}(\mathcal{D}_{train})$ using $(MLM)$.}
   Evaluate: $Perplexity \leftarrow  \mathcal{LM}$ on $\mathcal{D}_{test}$
 
 \caption{Spoken Language Modelling}\label{alg:alg1} 
\end{algorithm}

\begin{algorithm}
\SetKwInOut{Input}{Input}
\SetKwInOut{Output}{Output}
\Input{$\mathcal{P}_{Si}$, $\mathcal{T}_i$, $\mathcal{LM}$ from \autoref{alg:alg1}}
\Output{$(CONF_p,MSE)$}
 $(\mathcal{D}^{labelled}_{train}, \mathcal{D}^{labelled}_{dev}, \mathcal{D}^{labelled}_{test}) \leftarrow$ (train, dev, test set) according to  ($\mathcal{P}_{Si}$,$\mathcal{T}_i$)

 $CONF_p \leftarrow \mathcal{LM} + MLP $ 
 
  $CONF_p \leftarrow CONF_p(\mathcal{D}^{labelled}_{train})$ using $(MSE)$.
  
  Evaluate: $MSE \leftarrow  CONF_p$ on $\mathcal{D}_{test}$
 \caption{Confidence prediction}\label{alg:alg2}
\end{algorithm}


\textbf{Example of token representation strategies: }Our token representation strategies are built on the tokenizer introduced by \citet{devlin2018bert} and used the Sentence Piece algorithm \cite{kudo2018sentencepiece}. An example is given in \autoref{table:tokenizer_strat_additional}. 

\subsection{Dataset: Additional details}
We highlight relevant information about the dataset in \autoref{table:desc_stat}. The count of each ``uh" and ``um" filler is roughly the same. After discarding some videos due to missing labels, only 100 of them do not contain fillers. We use the original standard training, testing and validation folds provided in the CMU-Multimodal SDK \cite{SDK}. 

The process of transcription of fillers is described in \cite{park2014computational}.
The transcriptions were carried out via Amazon Mechanical Turk, using 18 native English speaking workers based in the United States. These workers were from the same pool of workers used to annotate the videos for high level attributes. Each transcription was then reviewed and edited by in-house experienced transcribers for accuracy.

\begin{table}[ht!]
\centering
\resizebox{\columnwidth}{!}{\begin{tabular}{l|l}
\hline
\textbf{Description} & \textbf{Value} \\ \hline
Videos that contain fillers & \textbf{792} \\ \hline
Total \textit{um} fillers in the corpus & 4969 \\ \hline
Total \textit{uh} fillers in the corpus & 4967 \\ \hline
Total fillers in the corpus & \textbf{9936} \\ \hline
Number of tokens in the corpus & 230462 \\ \hline
\% of tokens that are fillers & \textbf{4.31} \\ \hline
Average length (in tokens) of a video & 255.9 \\ \hline
\end{tabular}}
\caption{Details about the POM dataset}
\label{table:desc_stat}
\end{table}



In \autoref{table:sample} we give example sentences extracted from the POM dataset. In these examples, we can observe that the fillers are commonly located sentence-initially. Note, the corpus annotates ``uh" and ``um" as ``uhh" and ``umm" respectively, reflected in our examples taken from the dataset.

\begin{table}[ht]
    \centering
\resizebox{\columnwidth}{!}{\begin{tabular}{c} \hline
Samples \\\hline
\textbf{(umm)} the title actually translates to The Brotherhood of War. \\
\textbf{(umm)} The movie itself is a lot like Saving Private Ryan and Band of Brothers. \\
\textbf{(uhh)} Morgan Freeman is great in this movie, and \textbf{(uhh)} so is Tim Robbins. \\
\textbf{(umm)} You'll only like it if you're into kid of strange, bizarre humor. \\
It's just \textbf{(uhh)} pretty obvious stuff you know.\\
But \textbf{(umm)} a lot of the movie didn't really make sense.  \\
\textbf{(umm)} It's really funny, there there's (stutter) some really funny parts in it. \\
\textbf{(umm)} But, I recommend watching this movie it's really good. \\
\textbf{(umm)} The acting is only so-so. \\
And so \textbf{(umm)} I wouldn't really recommend it.\\
\textbf{(umm)} Yeah, but that's it. \\
\end{tabular}}
\caption{Some samples from the dataset. As can be seen, many of the fillers occur sentence-initially.}
\label{table:sample}
\end{table}

\subsection{Hyper-parameters for our experiments}
All the hyper-parameters have been optimised on the validation set based on the minimum of the training loss (MSE for confidence/sentiment prediction and perplexity for LM) accuracy computed on the last tag of the sequence.  We used Adam optimizer \cite{adam} with a learning rate of $10^-5$, which is updated using a polynomial decay. The gradient norm is clipped to 5.0, weight decay is set to $10^{-6}$, and dropout \cite{dropout} is set to 0.2. Models have been implemented in PyTorch and trained on a v100 using the same procedure as in \cite{colombo-etal-2019-affect,colombo2020guiding,witon-etal-2018-disney}.
\end{document}